\documentclass{article}

\usepackage[nonatbib, preprint]{neurips_2019}

\usepackage[utf8]{inputenc} % allow utf-8 input
\usepackage[T1]{fontenc}    % use 8-bit T1 fonts
\usepackage{hyperref}       % hyperlinks
\usepackage{url}            % simple URL typesetting
\usepackage{booktabs}       % professional-quality tables
\usepackage{amsfonts}       % blackboard math symbols
\usepackage{nicefrac}       % compact symbols for 1/2, etc.
\usepackage{microtype}      % microtypography
\usepackage{graphicx}
\usepackage{amsmath}
\usepackage{multirow}

\title{Decoupling feature propagation from the design of graph auto-encoders}

\author{%
   Paul M. Scherer \\ Dept. of Computer Science and Technology \\ University of Cambridge, UK 
   \And
   Helena Andres-Terre \\ Dept. of Computer Science and Technology \\ University of Cambridge, UK
   \And
   Pietro Li\'o \\ Dept. of Computer Science and Technology \\ University of Cambridge, UK
   \And
   Mateja Jamnik \\ Dept. of Computer Science and Technology \\ University of Cambridge, UK
}

\begin{document}

\maketitle

\begin{abstract}
We present two instances, L-GAE and L-VGAE, of the variational graph auto-encoding family (VGAE) based on separating feature propagation operations from graph convolution layers typically found in graph learning methods to a single linear matrix computation made prior to input in standard auto-encoder architectures. This decoupling enables the independent and fixed design of the auto-encoder without requiring additional GCN layers for every desired increase in the size of a node's local receptive field. Fixing the auto-encoder enables a fairer assessment on the size of a nodes receptive field in building representations. Furthermore a by-product of fixing the auto-encoder design often results in substantially smaller networks than their VGAE counterparts especially as we increase the number of feature propagations. A comparative downstream evaluation on link prediction tasks show comparable state of the art performance to similar VGAE arrangements despite considerable simplification. We also show the simple application of our methodology to more challenging representation learning scenarios such as spatio-temporal graph representation learning.

\end{abstract}

\section{Introduction}

Learning useful and efficient data representations is an important facet of machine learning research as they can heavily sway the performance of downstream pattern recognition and inference algorithms \cite{representationlearning}. Machine learning methods on graph structured data rely on the fundamental assumption that nodes are better contextualised and thus better represented through the utilisation of local relational information \cite{goyal, geometricdeeplearning, relationalinductivebias, representationlearninggraphs}. Hence, graph representation learning methods should also be designed to incorporate inductive biases to favour representations contextualising nodes based on its local receptive field \cite{HUBEL1959}.

Driven by the successful interpretation of this inductive bias through convolutional methods, Kipf and Welling's graph convolution layer (GCN) \cite{gcn} forms the basis of many current neural network designs on graph structured data. The variational graph auto-encoder (VGAE) and graph auto-encoder (GAE) \cite{vgae} also rely on GCN layers in its encoder followed by an inner product decoder of latent variables to reconstruct an adjacency matrix. A consequence of this design choice is that the number of layers in this network is directly linked to the size of a nodes local receptive field or the number of feature propagations performed on the features across the nodes in the graph.

We provide a simplification of the design by decoupling the feature propagation steps out of the auto-encoder into a single linear matrix multiplication performed on the dataset prior to input into the learning mechanism \cite{sgc}. This decouples the convolution operation from the layers in the auto-encoder allowing the design of the auto-encoder to be fixed regardless of the number of feature propagations we wish to perform. This allows the independent consideration of size of a nodes local receptive field and the application of standard off-the-shelf auto-encoder designs onto graph structured datasets. We call the resulting graph auto-encoder and variational graph auto-encoder L-GAE and L-VGAE respectively.

\section{Graphs and graph methods}

A great number of real world datasets exhibit some form of graph structure on top of their observations such as social networks, citation networks, protein-protein interaction networks to name a few. For example, within the Cora citation dataset \cite{cora} a node represents a scientific publication associated with a bag-of-words feature vector describing some of its content. Publications which have cited each other would be connected by an edge thereby the entire dataset forms a graph. The standard task is to then classify each of the nodes into one of 7 classes. Whilst it is wholly possible to apply neural networks on the node features themselves, disregarding the relationships between the nodes to perform the classification, empirical results have shown that classification performance was significantly better when the relations between nodes were taken into account. Hence extending algorithms to appropriately incorporate relational information in its learning process has attracted a great deal of attention \cite{goyal, geometricdeeplearning, relationalinductivebias, representationlearninggraphs}.

We define $\mathcal{G} = (V, E)$ as a graph where $V$ is a set of nodes and $E \subseteq (V \times V)$ be a 2-tuple set of edges in the graph. Hence if $u$ and $v$ are nodes in $\mathcal{G}$, their relation would be recorded with an edge as $(u, v) \in E$. The \textit{neighbours} of a node $v$ in graph $\mathcal{G}=(V,E)$,  is the set of nodes which share an edge with $v$, denoted $\mathcal{N}(v) = \{u | (v,u) \in E \}$. 

For a graph with $n$ nodes, $\mathbf{A} \in \mathbb{R}^{n \times n}$ is a symmetric adjacency matrix where element $a_{i,j}$ is the weight of the edge between nodes $v_i$ and $v_j$. If $(v_i, v_j) \notin E$ then $A_{i,j} = 0$. A diagonal degree matrix $\mathbf{D} \in \mathbb{R}^{n \times n}$ is defined as the matrix where each entry on the diagonal is the row-sum of the adjacency matrix. For graphs with node features, each node $v_i$ has an associated $d$-dimensional feature vector $\mathbf{x}_i \in \mathbb{R}^d$. The feature matrix $\mathbf{X} \in \mathbb{R}^{n \times d}$ represents feature vectors for every node in the graph.

\subsection{Graph convolutional networks (GCN)} \label{section: gcn}
Graph convolutional networks learn feature representations for nodes through several iterations of graph convolution layers in a composite manner. At each layer node representations $\mathbf{x}_i$ are updated through three stages: localized feature propagation, linear transformation, and a element-wise non-linear activation \cite{sgc}. The node representations fed into the $k$-th layer of a GCN can be denoted as $\mathbf{H}^{k-1}$ with its output being $\mathbf{H}^{k}$, initially $\mathbf{H}^{0} = X$ will be the input to the first GCN layer.

In the feature propagation stage of a GCN layer the features $\mathbf{h}_i$ of each node $v_i$ is averaged with the feature vectors of its neighbours $\mathcal{N}(v_i)$.

\begin{equation} \label{eq: featureprop}
\bar{\mathbf{h}}_{i}^{(k)} = \frac{1}{d_i + 1}\mathbf{h}_{i}^{(k-1)} + \sum_{j=1}^{n} \frac{a_{i,j}}{\sqrt{(d_i + 1)(d_j + 1)}}\mathbf{h}_{k}^{(k-1)}
\end{equation}

This can be expressed as the matrix operation by finding a normalized adjacency matrix $\mathbf{S}$ with added self loops defined as $\mathbf{S} = \tilde{\mathbf{D}}^{-\frac{1}{2}} \tilde{\mathbf{A}} \tilde{\mathbf{D}}^{-\frac{1}{2}}$ where $\tilde{\mathbf{A}} = \mathbf{A}+\mathbf{I}$ and $\tilde{\mathbf{D}} = \mathbf{D} + \mathbf{I}$ (the renormalization trick \cite{gcn}). Then equation \ref{eq: featureprop} can be summarized as:

\begin{equation}\label{eq: featurepropmatmul}
\bar{\mathbf{H}}^{(k)} = \mathbf{SH}^{(k-1)}
\end{equation}

% Intuitively, this step smoothes the hidden representations localy along the edges of the graph and ultimately encourages similar predictions among locally connected nodes

After the propagation of the node features in local neighbourhoods the rest of GCN layer is similar to a standard feed forward multi-layer perceptron network (MLP). The hidden representations $\bar{\mathbf{H}}^{(k)}$ at the $k$-th layer in multiplied with a weight matrix $\mathbf{W}^{(k)}$ and passed into a point-wise non-linear activation such as the ReLU function \cite{relu}.

\begin{equation} \label{eq: gcnactivation}
\mathbf{H}^{(k)} = \text{ReLU}(\bar{\mathbf{H}}^{(k)} \mathbf{W}^{(k)})
\end{equation}

\subsection{Variational graph auto-encoders} \label{sec: vgae}

Kipf and Welling's VGAE \cite{vgae} was born out of the combination of graph convolutional layers \cite{gcn} and the variational auto-encoder \cite{vae}. Focusing on link prediction as a target task, the VGAE is designed to capture the higher order dependencies between nodes using the adjacency matrix $\mathbf{A}$ as part of its input to the model and reconstructing the matrix at the decoder using an inner product of the latent variables.

The probabilistic encoder learns the distribution $Q(\mathbf{Z} | \mathbf{X,A})$ that is defined as two graph convolutional layers as defined in equation \ref{eq: featurepropmatmul}.

\begin{equation} \label{eq: vgaeencoders}
Q(\mathbf{Z} | \mathbf{X,A}) = \prod_{i=1}^{N} Q(\mathbf{z}_i | \mathbf{X,A}), \thinspace \textrm{with} \thinspace \thinspace \thinspace Q(\mathbf{z}_i | \mathbf{X,A}) =  \mathcal{N}(\mathbf{z}_i | \mathbf{\mu}_i, \textrm{diag}(\mathbf{\sigma}_{i}^{2}))
\end{equation}

Here $\mathbf{\mu}_i \in \mathbf{M}$ and $\mathbf{M} = \textrm{GCN}_{\mathbf{\mu}}(\mathbf{X,A})$ is the matrix of mean vectors $\mathbf{\mu}_i$. Following on, $\log \mathbf{\sigma} = \textrm{GCN}_{\mathbf{\sigma}}(\mathbf{X,A})$. The two layer GCN encoder is thus defined as $\textrm{GCN} = \mathbf{S}(\textrm{ReLU}(\mathbf{SXW}_0))\mathbf{W}_1$. The probabilistic decoder is then defined as a single inner product of the latent variables

\begin{equation} \label{eq: VGAE_DECODER}
p(\mathbf{A} | \mathbf{Z}) = \prod_{i=1}^{N} \prod_{j=1}^{N} p(A_{i,j} | \mathbf{z}_i, \mathbf{z}_j), \thinspace \textrm{with} \thinspace \thinspace \thinspace p(A_{i,j} | \mathbf{z}_i, \mathbf{z}_j) = \sigma(\mathbf{z}_{i}^{T} \mathbf{z}_j)
\end{equation}

where $\sigma(\cdot)$ is an element-wise sigmoid function or more simply expressed as a matrix product $\hat{\mathbf{A}} = \sigma(\mathbf{Z}^{T}\mathbf{Z})$ wherein $\hat{\mathbf{A}}$ is interpreted as the networks' reconstructed probabilistic adjacency matrix. The variational lower bound $\mathcal{L}$ is optimized with respect to the networks parameters $\mathbf{W}$

\begin{equation} \label{eq: LossVGAE}
\mathcal{L} = \mathbb{E}_{Q(\mathbf{Z} | \mathbf{X,A})}[\log P(\mathbf{A}|\mathbf{Z})] - KL[Q(\mathbf{Z} | \mathbf{X,A}) || P(\mathbf{Z})]
\end{equation}

One implication of using GCN layers in the design of these graph auto-encoders is that the number of layers used in the encoder is directly linked to the number of feature propagations performed upon the features of the graph. Within the VGAE encoder we see this happen twice, which is interpreted as two stages of feature propagation amongst the nodes of the graph through multiplication with $\mathbf{S}$. To integrate information from larger neighbourhoods per node, another GCN layer would have to be appended to the encoder changing the architecture of the model.

\section{L-GAE and L-VGAE}
The recursive application non-linear activation and learned weight matrices onto the propagated features shown in equation \ref{eq: gcnactivation} creates considerable overhead for large graphs \cite{sgc}. This overhead may not be critical as previous works have shown that much of the power network based feature classifiers gain comes from local feature propagation \cite{labelprop, labelprop2, sgc}. Wu et al. \cite{sgc} have hence suggested removing all non-linear transitions except a final softmax layer for classification. We also remove this activation and weights to define a $k$-hop smoothed node feature matrix using $\mathbf{S}$ as defined in section \ref{section: gcn}.

\begin{equation} \label{eq:xbar}
\begin{aligned}
\bar{\mathbf{X}} = \mathbf{SS...SX} \\
\bar{\mathbf{X}} = \mathbf{S}^{k}\mathbf{X}
\end{aligned}
\end{equation}

$\bar{\mathbf{X}}$ can be interpreted as a parameter-free feature preprocessing step on the graph. This reduces any downstream node feature task such as classification, regression, clustering, etc. on training and evaluating standard vector models on the pre-processed $\bar{\mathbf{X}}$. Each application of $\mathbf{S}$ performs a localized feature propagation across the graph as defined in equation \ref{eq: featureprop}. This considerably simplifies the construction of graph auto-encoders. As an example, a L-VGAE variational graph auto-encoder which considers a 3-hop local receptive field for each node can be constructed by feeding $\bar{\mathbf{X}} = \mathbf{S}^3\mathbf{X}$ into any variational auto-encoder.

\section{Evaluation and discussion}
\begin{table}[]
\centering
\caption{Link prediction task in citation networks recorded with mean and standard deviation. The (*) signifies instances where the model was not given feature information for each node.}
\begin{tabular}{@{}lllllll@{}}
\toprule
\multirow{2}{*}{\textbf{Method}} & \multicolumn{2}{c}{\textbf{Cora}}                & \multicolumn{2}{c}{\textbf{Citeseer}}            & \multicolumn{2}{c}{\textbf{PubMed}}              \\
                                 & \multicolumn{1}{c}{AUC} & \multicolumn{1}{c}{AP} & \multicolumn{1}{c}{AUC} & \multicolumn{1}{c}{AP} & \multicolumn{1}{c}{AUC} & \multicolumn{1}{c}{AP} \\ \midrule
SC                               & 84.6 $\pm$ 0.01             & 88.5 $\pm$ 0.00            & 80.5 $\pm$ 0.01             & 85.0 $\pm$ 0.01            & 84.2  $\pm$ 0.02            & 87.8 $\pm$ 0.01            \\
DW                               & 83.1 $\pm$ 0.01             & 85.0 $\pm$ 0.00            & 80.5 $\pm$ 0.02             & 83.6 $\pm$ 0.01            & 84.4 $\pm$ 0.00             & 84.1 $\pm$ 0.00            \\ \midrule
GAE*                             & 84.3 $\pm$ 0.02             & 88.1 $\pm$ 0.01            & 78.7 $\pm$ 0.02             & 84.1 $\pm$ 0.02            & 82.2 $\pm$ 0.01             & 87.4 $\pm$ 0.00            \\
VGAE*                            & 84.0 $\pm$ 0.02             & 87.7 $\pm$ 0.01            & 78.9 $\pm$ 0.03             & 84.1 $\pm$ 0.02            & 82.7 $\pm$ 0.01             & 87.5 $\pm$ 0.01            \\
L-GAE*                           & \textbf{86.3 $\pm$ 0.01}                      & \textbf{89.0 $\pm$ 0.01}                     & \textbf{80.4 $\pm$ 0.01}                      & \textbf{84.2 $\pm$ 0.01} & 82.4 $\pm$ 0.00                      & 87.5 $\pm$ 0.00                     \\
L-VGAE*                          & 85.9 $\pm$ 0.01                      & 88.2 $\pm$ 0.01                     & 79.8  $\pm$ 0.00                    &  83.9 $\pm$ 0.00                    & \textbf{83.1 $\pm$ 0.01}   &   \textbf{87.6 $\pm$ 0.01                    } \\ \midrule
GAE                              & 91.0 $\pm$ 0.02             & 92.0 $\pm$ 0.03            & 89.5 $\pm$ 0.04             & 89.9 $\pm$ 0.05            & \textbf{96.4 $\pm$ 0.00}             & \textbf{96.5 $\pm$ 0.00} \\
VGAE                             & 91.4 $\pm$ 0.01             & 92.6 $\pm$ 0.01            & 90.8 $\pm$ 0.02             & 92.0 $\pm$ 0.02            & 94.4 $\pm$ 0.02             & 94.7 $\pm$ 0.02            \\
L-GAE                            &  91.9 $\pm$ 0.00                     & 93.0 $\pm$ 0.00                     & 90.7 $\pm$ 0.01                      & 91.3 $\pm$ 0.01                     & 95.4 $\pm$ 0.00                  & 95.6 $\pm$ 0.00                      \\
L-VGAE                           & \textbf{92.9 $\pm$ 0.00  }                    & \textbf{93.9 $\pm$ 0.00 }                    & \textbf{92.5 $\pm$ 0.00}                      & \textbf{93.6 $\pm$ 0.00}                &   92.6 $\pm$ 0.01                 & 92.8 $\pm$ 0.01                     \\ \bottomrule
\end{tabular}
\label{tab: results}
\end{table}

A comparative evaluation was performed on the network reconstruction task set by VGAE \cite{vgae}. For a "fair" comparison between the GAE/VGAE and the proposed L-GAE/L-VGAE in the benchmark tests the design of the networks was made to be as similar as possible. For GAE/VGAE we adopt the definition made in their publication \cite{vgae}: a two GCN layer encoder with 32 and 16 hidden units incorporating 2 feature propagations and the inner product decoder (equation \ref{eq: VGAE_DECODER}). In our methodology this corresponds to 2 feature propagation on the feature matrix, i.e. $k=2$ in equation \ref{eq:xbar}, followed by an encoder of two linear layers with 32 and 16 hidden units and the same inner product decoder. Each of the presented models was trained for 200 epochs using the same Adam optimizer. 

Experiments were run on two streams for the auto-encoder models. One stream did not use feature vector for each node in the graph, implemented by $\mathbf{X} = \mathbf{I}$, and one stream of results utilised the feature vectors for each node. The results shown in table \ref{tab: results} show that L-GAE and L-VGAE perform on par and outperform their GAE and VGAE counterparts in both tasks. We discuss the by-product of smaller networks in the supplementary material.

% \usepackage{booktabs}
\begin{table}[]
\centering
\caption{Regression performance over time on METR-LA with mean absolute error score and standard deviation on the predictions of the baseline LSTM model applied on raw features as well the baseline LSTM model applied on features learned with L-VGAE and STDGI.}
\label{tab: resultsST}
\begin{tabular}{@{}llll@{}}
\toprule
\textbf{}         & \multicolumn{3}{c}{\textbf{Mean absolute error}}                                               \\
\textbf{Method}   & 15 min                   & 30 min                   & 60 min                   \\ \midrule
LSTM Baseline     & 3.67 $\pm$ 0.04          & 4.88 $\pm$ 0.02          & 6.53 $\pm$ 0.02          \\
STDGI \cite{stdgi}            & 3.59 $\pm$ 0.00          & \textbf{4.78 $\pm$ 0.00} & \textbf{6.34 $\pm$ 0.01} \\
L-VGAE + Baseline & \textbf{3.58 $\pm$ 0.00} & 4.84 $\pm$ 0.01          & 6.43 $\pm$ 0.01          \\ \bottomrule
\end{tabular}
\end{table}

%\section{Conclusion}
We have not proposed a model that is especially innovative. A fixed low-pass filter for graphs was studied very recently in Wu et al. \cite{sgc} but we presented a simple methodology to design graph auto-encoders by decoupling the feature propagations. This simplification allows us to quickly build methods for more challenging scenarios such as unsupervised learning of spatio-temporal node representations as covered in Opolka et al. \cite{stdgi} on the METR-LA dataset \cite{metr-la}. In this scenario we have constructed a simple non-forward looking model through simple application of $\bar{\mathbf{X}}$ into an Encoder-Decoder LSTM \cite{seq2seq} and achieved an improvement over the baseline supervised method which did not take relations into account. We are on par with the STDGI in short time horizons, however do worse as the later time predictions which makes sense as we do not incorporate future states as part of learning as in STDGI \cite{stdgi} and only relational information. We could potentially improve this by using a more sophisticated temporal model such as a flow-based generative model based around WaveNet \cite{wavenet} or a temporal auto-encoder. We will continue expanding on this work towards other challenging scenarios and hope that this could be seen as a "first step" or baseline method to consider when designing unsupervised graph representation techniques.

\bibliographystyle{unsrt} 
\bibliography{neurips_2019} 
\newpage
\appendix

\section{Appendix A: potentially smaller network in large receptive fields}

An interesting aspect of auto-encoders is to look at the size and number of trainable parameters, or degrees of freedom, present in the model architectures. As the design of the auto-encoder in the GAE/VGAE is intimately coupled with the desired number of feature propagations, the resulting \textit{size} of the network gets considerably larger with each additional feature propagation. In L-GAE/L-VGAE models the feature propagation of features is processed independently prior to input into the auto-encoder meaning that much smaller networks can be employed on the different levels of feature propagation. This enables the study of feature propagation depth as an independent variable on the same network and in model selection as an independent hyper-parameter. Tables \ref{cora_weights}, \ref{citeseer_weights}, \ref{pubmed_weights} show the number of degrees of freedom present in each auto-encoder based on number of feature propagation steps for Cora, Citeseer, and PubMed networks respectively. The network used in the L-GAE and L-VGAE are the same used in the evaluation in table 1, whilst the VGAE requires a new hidden layer with base 2 progression as implied in the original design of using 32 and 16 hidden units.

%%%%%
%% CORA
% \usepackage{booktabs}
\begin{table}[h]
\center
\caption{Number of trainable parameters in the auto-encoder given the number of feature propagations for the Cora dataset. Note that the size of the networks for L-GAE and L-VGAE are fixed, we chose a two linear layer encoder (32 and 16 hidden units each) and single layer inner product decoder to match the architecture of the encoder model described in Kipf and Welling. The auto-encoder in the VGAE has to include a layer for each feature propagation step, sizes used are based on base 2 progression as implied in Kipf and Welling. ie for $k=1$ (16 hidden unit layer in auto-encoder), $k=2$ (32, 16), $k=3$ (64, 32, 16), and $k=7$ (1024, 512, 256, 128, 64, 32, 16)}
\label{cora_weights}
\begin{tabular}{@{}lllll@{}}
\toprule
                & \multicolumn{4}{l}{\textbf{Number of Feature Propagation Steps}}                             \\
\textbf{Method} & k = 1 & \begin{tabular}[c]{@{}l@{}}k = 2\\ (Base Configuration)\end{tabular} & k = 3 & k = 7 \\ \midrule
L-GAE           & 46416   & \textbf{46416}                                                                  & \textbf{46416}   & \textbf{46416}   \\
L-VGAE          & 46944   & \textbf{46944}                                                                  & \textbf{46944}   & \textbf{46944}   \\
VGAE            & \textbf{45856}   & 48928                                                                  & 94784   & 2166784   \\ \bottomrule
\end{tabular}
\end{table}

%%%%%
%% CiteSeer
% \usepackage{booktabs}
\begin{table}[h]
\center
\caption{Number of trainable parameters in the auto-encoder given the number of feature propagations for the CiteSeer dataset.}
\label{citeseer_weights}
\begin{tabular}{@{}lllll@{}}
\toprule
                & \multicolumn{4}{l}{\textbf{Number of Feature Propagation Steps}}                             \\
\textbf{Method} & k = 1 & \begin{tabular}[c]{@{}l@{}}k = 2\\ (Base Configuration)\end{tabular} & k = 3 & k = 7 \\ \midrule
L-GAE           & 119056   & \textbf{119056}                                                                  & \textbf{119056}   & \textbf{119056}   \\
L-VGAE          & 119584   & \textbf{119584}                                                                  & \textbf{119584}   & \textbf{119584}   \\
VGAE            & \textbf{118496}   & 121568                                                                  & 240064   & 4491264   \\ \bottomrule
\end{tabular}
\end{table}

%%%%%
%% PubMed
% \usepackage{booktabs}
\begin{table}[h]
\center
\caption{Number of trainable parameters in the auto-encoder given the number of feature propagations for the Pubmed dataset.}
\label{pubmed_weights}
\begin{tabular}{@{}lllll@{}}
\toprule
                & \multicolumn{4}{l}{\textbf{Number of Feature Propagation Steps}}                             \\
\textbf{Method} & k = 1 & \begin{tabular}[c]{@{}l@{}}k = 2\\ (Base Configuration)\end{tabular} & k = 3 & k = 7 \\ \midrule
L-GAE           & 16560   & \textbf{16560}                                                                  & \textbf{16560}   & \textbf{16560}   \\
L-VGAE          & 17088   & \textbf{17088}                                                                  & \textbf{17088}   & \textbf{17088}   \\
VGAE            & \textbf{16000}   & 19072                                                                  & 35072   & 1211392   \\ \bottomrule
\end{tabular}
\end{table}

\end{document}